\documentstyle[subfloat,treemacros,twoside,cl]{article}

\issue{24}{4}{1998}

\title{Selective Sampling for Example-based Word Sense Disambiguation}

\author{
  Atsushi Fujii\thanks{Department of Library and Information
    Science, University of Library and Information Science, 1-2
    Kasuga, Tsukuba, 305-8550, Japan} &
  Kentaro Inui\thanks{Department of Artificial Intelligence, Faculty
    of Computer Science and Systems Engineering, Kyushu Institute of
    Technology, 680-4, Kawazu, Iizuka, Fukuoka 820-0067, Japan} \\
  \affil{University of Library and Information Science} &
  \affil{Kyushu Institute of Technology} \\ \\
  Takenobu Tokunaga\thanks{Department of Computer Science, Tokyo Institute
    of Technology, 2-12-1 Oookayama Meguroku Tokyo 152-8552, Japan} &
  Hozumi Tanaka$^{\tiny \fnsymbol{footnote}}$ \\
  \affil{Tokyo Institute of Technology} &
  \affil{Tokyo Institute of Technology}}

\runningtitle{Selective Sampling}
\runningauthor{Fujii, Inui, Tokunaga, and Tanaka}

\renewcommand{\footmark}{\copyright\ 1998 Association for
  Computational Linguistics}

\input{psfig}

\begin{document}
\maketitle

\newcommand{\E}{\mbox{$e$}}
\newcommand{\EX}[2]{\mbox{${\cal E}_{#1}^{#2}$}}
\newcommand{\EXi}[2]{\mbox{${\cal E}_{#1,#2}$}}
\newcommand{\EXii}[3]{\mbox{${\cal E}^{#1}_{#2,#3}$}}
\newcommand{\V}{\mbox{$v$}}
\newcommand{\SS}{\mbox{$s$}}
\newcommand{\SSi}[1]{\mbox{$s_{#1}$}}
\newcommand{\C}{\mbox{$c$}}
\newcommand{\Ci}[1]{\mbox{$c_{#1}$}}
\newcommand{\Ni}[1]{\mbox{$n_{#1}$}}
\newcommand{\Mi}[1]{\mbox{$m_{#1}$}}
\newcommand{\X}[1]{\mbox{${\cal X}_{#1}$}}
\newcommand{\eq}[1]{(\ref{#1})}
\newcommand{\set}[1]{\mbox{\bf #1}}
\newcommand{\seti}[2]{\mbox{$\set{#1}_{#2}$}}
\newcommand{\colcenter}[1]{\hfill\centering #1\hfill}
\newcommand{\etal}{et~al.}
\newcommand{\etaleos}{et~al}

\begin{abstract}
  This paper proposes an efficient example sampling method for
  example-based word sense disambiguation systems. To construct a
  database of practical size, a considerable overhead for manual sense
  disambiguation (overhead for supervision) is required. In addition,
  the time complexity of searching a large-sized database poses a
  considerable problem (overhead for search). To counter these
  problems, our method selectively samples a smaller-sized effective
  subset from a given example set for use in word sense
  disambiguation.  Our method is characterized by the reliance on the
  notion of training utility: the degree to which each example is
  informative for future example sampling when used for the training
  of the system. The system progressively collects examples by
  selecting those with greatest utility.  The paper reports the
  effectiveness of our method through experiments on about one thousand
  sentences. Compared to experiments with other example sampling
  methods, our method reduced both the overhead for supervision and
  the overhead for search, without the degeneration of the performance
  of the system.
\end{abstract}

\section{Introduction}
\label{sec:intro}

Word sense disambiguation is a potentially crucial task in many NLP
applications, such as machine translation~\cite{brown:acl-91},
parsing~\cite{lytinen:aaai-86,k.nagao:ieice-94} and text
retrieval~\cite{krovets:acmtois-92,voorhees:sigir-93}. Various
corpus-based approaches to word sense disambiguation have been
proposed~\cite{bruce:acl-94,charniak:93,dagan:cl-94,fujii:coling-96,hearst:oed-91,karov:wvlc-96,kurohashi:ieice-94,x.li:ijcai-95,ng:acl-96,niwa:coling-94,schutze:supercomp-92,uramoto:ieice-94,yarowsky:acl-95}.
The use of corpus-based approaches has grown with the use of
machine-readable texts, because unlike conventional rule-based
approaches relying on hand-crafted selectional rules (some of which
are reviewed, for example, by Hirst~\shortcite{hirst:87}),
corpus-based approaches release us from the task of generalizing
observed phenomena through a set of rules. Our verb sense
disambiguation system is based on such an approach, that is, an
example-based approach.  A preliminary experiment showed that our
system performs well when compared with systems based on other
approaches, and motivated us to further explore the example-based
approach (we elaborate on this experiment in
Section~\ref{subsec:pre_experiment}). At the same time, we concede
that other approaches for word sense disambiguation are worth further
exploration, and while we focus on example-based approach in this
paper, we do not wish to draw any premature conclusions regarding the
relative merits of different generalized approaches.

As with most example-based
systems~\cite{fujii:coling-96,kurohashi:ieice-94,x.li:ijcai-95,uramoto:ieice-94},
our system uses an example-database (database, hereafter) which
contains example sentences associated with each verb sense. Given an
input sentence containing a polysemous verb, the system chooses the
most plausible verb sense from predefined candidates.  In this
process, the system computes a scored similarity between the input and
examples in the database, and chooses the verb sense associated with
the example which maximizes the score.  To realize this, we have to
manually disambiguate polysemous verbs appearing in examples, prior to
their use by the system. We shall call these examples {\bf supervised
examples}.  A preliminary experiment on eleven polysemous Japanese
verbs showed that (a) the more supervised examples we provided to the
system, the better it performed, and (b) in order to achieve a
reasonable result (say over 80\% accuracy), the system needed a
hundred-order supervised example set for each verb. Therefore, in
order to build an operational system, the following problems have to
be taken into account\footnote{Note that these problems are associated
with corpus-based approaches in general, and have been identified by a
number of
researchers~\cite{engelson:acl-96,lewis:sigir-94,uramoto:coling-94,yarowsky:acl-95}.}:
\begin{itemize}
\item given human resource limitations, it is not reasonable to
  supervise every example in large corpora (``overhead for
  supervision''),
\item given the fact that example-based systems, including our system,
  search the database for the examples most similar to the input, the
  computational cost becomes prohibitive if one works with a very
  large database size (``overhead for search'').
\end{itemize}
These problems suggest a different approach, namely to {\it select\/}
a small number of optimally informative examples from given corpora. 
Hereafter we will call these examples {\bf samples}.

\begin{figure}
  \centering
  \mbox{\psfig{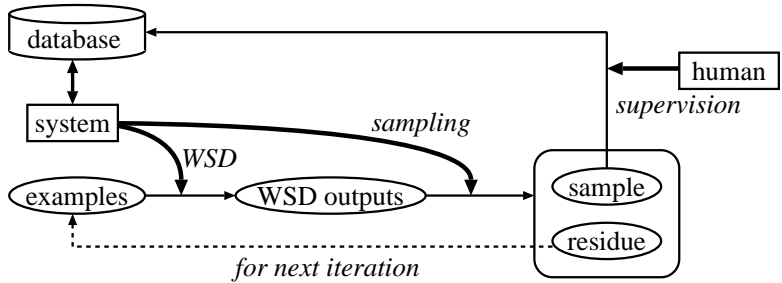}}
  \medskip
  \caption{Flow of control of the example sampling system.}
  \label{fig:concept}
\end{figure}

Our example sampling method, based on the utility maximization
principle, decides on the preference for including a given example in
the database.  This decision procedure is usually called {\bf
selective sampling}~\cite{cohn:ml-94}. The overall control flow of
selective sampling systems can be depicted as in
Figure~\ref{fig:concept}, where ``system'' refers to our verb sense
disambiguation system, and ``examples'' refers to an unsupervised
example set.  The sampling process basically cycles between the word
sense disambiguation (WSD) and training phases. During the WSD phase,
the system generates an interpretation for each polysemous verb
contained in the input example (``WSD outputs'').  This phase is
equivalent to normal word sense disambiguation execution.  During the
training phase, the system selects samples for training from the
previously produced outputs. During this phase, a human expert
supervises samples, that is, provides the correct interpretation for
the verbs appearing in the samples. Thereafter, samples are simply
incorporated into the database without any computational overhead (as
would be associated with globally reestimating parameters in
statistics-based systems), meaning that the system can be trained on
the remaining examples (the ``residue'') for the next
iteration. Iterating between these two phases, the system
progressively enhances the database. Note that the selective sampling
procedure gives us an optimally informative database of a given size
irrespective of the stage at which processing is terminated.

Several researchers have proposed this type of approach for NLP
applications.  Engelson and Dagan~\shortcite{engelson:acl-96} proposed
a committee-based sampling method, which is currently applied to HMM
training for part-of-speech tagging. This method sets several models
(the committee) taken from a given supervised data set, and selects
samples based on the degree of disagreement among the committee
members as to the output. This method is implemented for
statistics-based models.  However, to formalize and map the concept of
selective sampling into example-based approaches has yet to be
explored.

Lewis and Gale~\shortcite{lewis:sigir-94} proposed an uncertainty
sampling method for statistics-based text classification. In this
method, the system always samples outputs with an uncertain level of
correctness. In an example-based approach, we should take into account
the training effect a given example has on other unsupervised
examples. This is introduced as {\bf training utility} in our method.
We devote Section~\ref{sec:eval} to further comparison of our approach
and other related works.

With respect to the problem of overhead for search, possible solutions
would include the generalization of similar
examples~\cite{kaji:coling-92,nomiyama:tipsj-93} or the reconstruction
of the database using a small portion of useful instances selected
from a given supervised example set~\cite{aha:ml-91,smyth:ijcai-95}.
However, such approaches imply a significant overhead for supervision
of each example prior to the system's execution. This shortcoming is
precisely what our approach aims to avoid: we aim to reduce the
overhead for supervision as well as the overhead for search.

Section~\ref{sec:vader} describes the basis of our verb sense
disambiguation system and preliminary experiment, in which we compared
our method with other disambiguation methods. 
Section~\ref{sec:sampling} then elaborates on our example sampling
method. Section~\ref{sec:eval} reports on the results of our
experiments through comparison with other proposed selective sampling
methods, and discusses theoretical differences between those methods.

\section{Example-Based Verb Sense Disambiguation System}
\label{sec:vader}

\subsection{The Basic Idea}
\label{subsec:basis}

Our verb sense disambiguation system is based on the method proposed
by Kurohashi and Nagao~\shortcite{kurohashi:ieice-94} and later
enhanced by Fujii~\etal~\shortcite{fujii:coling-96}.  The system uses
a database containing examples of collocations for each verb sense and
its associated case frame(s). Figure~\ref{fig:database} shows a
fragment of the entry associated with the Japanese verb {\it
toru}. The verb {\it toru\/} has multiple senses, a sample of which
are `to take/steal,' `to attain,' `to subscribe,' and `to reserve.'
The database specifies the case frame(s) associated with each verb
sense. In Japanese, a complement of a verb consists of a noun phrase
(case filler) and its case marker suffix, for example {\it ga\/}
(nominative) or {\it wo\/} (accusative). The database lists several
case filler examples for each case. The task of the system is to
``interpret'' the verbs occurring in the input text, i.e., to choose
one sense from among a set of candidates.\footnote{Note that unlike
the automatic acquisition of word sense
definitions~\cite{fukumoto:coling-94,pustejovsky:ai-93,utsuro:coling-96,zernik:ijcai-89},
the task of the system is to identify the best matched category with a
given input, from {\it predefined\/} candidates.} All verb senses we
use are defined in IPAL~\cite{ipa:87}, a machine-readable
dictionary. IPAL also contains example case fillers as shown in
Figure~\ref{fig:database}. Given an input, which is currently limited
to a simple sentence, the system identifies the verb sense on the
basis of the scored similarity between the input and the examples
given for each verb sense.  Let us take the sentence below as an
example input:
\begin{list}{}{\setlength{\leftmargin}{0mm}}
\item
  \begin{tabular}{lll}
    {\it hisho\/} {\it ga\/} & {\it shindaisha\/} {\it wo\/} & {\it
      toru}. \\ (secretary-NOM) & (sleeping car-ACC) & (?)
  \end{tabular}
\end{list}

In this example, one may consider {\it hisho\/} (`secretary') and {\it
shindaisha\/} (`sleeping car') to be semantically similar to {\it
joshu\/} (`assistant') and {\it hikouki\/} (`airplane') respectively,
and since both collocate with the `to reserve' sense of {\it toru},
one could infer that {\it toru\/} should be interpreted as `to
reserve.' This resolution originates from the analogy
principle~\cite{nagao:ahi-84}, and can be called nearest neighbor
resolution because the verb in the input is disambiguated by
superimposing the sense of the verb appearing in the example of
highest similarity.\footnote{In this paper, we use ``example-based
systems'' to refer to systems based on nearest neighbor resolution.}
The similarity between an input and an example is estimated based on
the similarity between case fillers marked with the same case.

\begin{figure}
  \scriptsize
  \leavevmode
  \begin{tabular}{|lll|} \hline
    & & \\
    \framebox{{\it toru\/}:} & & \\ & & \\ \hline
    $\left\{\begin{tabular}{ll}
        {\it suri\/} & (pickpocket) \\ {\it kanojo\/} & (she) \\
        {\it ani\/} & (brother) \end{tabular}\right\}$ {\it ga\/} &
    $\left\{\begin{tabular}{ll}
        {\it kane\/} & (money) \\ {\it saifu\/} & (wallet) \\
        {\it otoko\/} & (man) \\ {\it uma\/} & (horse) \\
        {\it aidea\/} & (idea) \end{tabular}\right\}$ {\it wo\/} &
    {\it toru\/} (to take/steal) \\ \hline
    $\left\{\begin{tabular}{ll}
        {\it kare\/} & (he) \\ {\it kanojo\/} & (she) \\
        {\it gakusei\/} & (student) \end{tabular}\right\}$ {\it ga\/} &
    $\left\{\begin{tabular}{ll}
        {\it menkyoshou\/} & (license) \\ {\it shikaku\/} & (qualification) \\
        {\it biza\/} & (visa) \end{tabular}\right\}$ {\it wo\/} &
    {\it toru\/} (to attain) \\ \hline
    $\left\{\begin{tabular}{ll}
        {\it kare\/} & (he) \\ {\it chichi\/} & (father) \\
        {\it kyaku\/} & (client) \end{tabular}\right\}$ {\it ga\/} &
    $\left\{\begin{tabular}{ll} {\it shinbun\/} & (newspaper) \\
        {\it zasshi\/} & (journal) \end{tabular}\right\}$ {\it wo\/} &
    {\it toru\/} (to subscribe) \\ \hline
    $\left\{\begin{tabular}{ll} {\it kare\/} & (he) \\
        {\it dantai\/} & (group) \\ {\it ryokoukyaku\/} & (passenger) \\
        {\it joshu\/} & (assistant) \end{tabular}\right\}$ {\it ga\/} &
    $\left\{\begin{tabular}{ll}
        {\it kippu\/} & (ticket) \\ {\it heya\/} & (room) \\
        {\it hikouki\/} & (airplane) \end{tabular}\right\}$ {\it wo\/} &
    {\it toru\/} (to reserve) \\ \hline
    {\hfill \centering $\vdots$ \hfill} & {\hfill \centering $\vdots$
      \hfill} & {\hfill \centering $\vdots$ \hfill} \\ \hline
  \end{tabular}
  \medskip
  \caption{A fragment of the database, and the entry
    associated with the Japanese verb {\it toru}.}
  \label{fig:database}
\end{figure}

Furthermore, since the restrictions imposed by the case fillers in
choosing the verb sense are not equally selective,
Fujii~\etal~\shortcite{fujii:coling-96} proposed a weighted case
contribution to disambiguation (CCD) of the verb senses.  This CCD
factor is taken into account when computing the score for each sense
of the verb in question.  Consider again the case of {\it toru} in
Figure~\ref{fig:database}. Since the semantic range of nouns
collocating with the verb in the nominative does not seem to have a
strong delinearization in a semantic sense (in
Figure~\ref{fig:database}, the nominative of each verb sense displays
the same general concept, i.e., {\footnotesize HUMAN}), it would be
difficult, or even risky, to properly interpret the verb sense based
on similarity in the nominative. In contrast, since the semantic
ranges are disparate in the accusative, it would be feasible to rely
more strongly on similarity here.

\begin{figure}
  \centering
  \mbox{\psfig{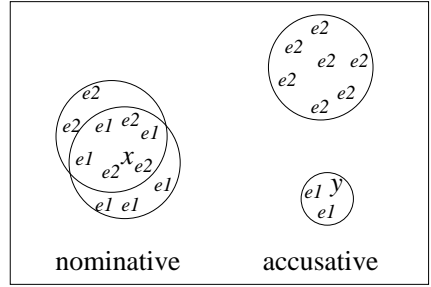}}
  \medskip
  \caption{The semantic ranges of the nominative and accusative for
    the verb {\it toru}.}
  \label{fig:ccd}
\end{figure}

This argument can be illustrated as in Figure~\ref{fig:ccd}, in which
the symbols $e_1$ and $e_2$ denote example case fillers of different
case frames, and an input sentence includes two case fillers denoted
by $x$ and $y$.  The figure shows the distribution of example case
fillers for the respective case frames, denoted in a semantic space.
The semantic similarity between two given case fillers is represented
by the physical distance between the two symbols.  In the nominative,
since $x$ happens to be much closer to an $e_2$ than any $e_1$, $x$
may be estimated to belong to the range of $e_2$'s, although $x$
actually belongs to both sets of $e_1$'s and $e_2$'s.  In the
accusative, however, $y$ would be properly estimated to belong to the
set of $e_1$'s due to the disjunction of the two accusative case
filler sets, even though examples do not fully cover each of the
ranges of $e_1$'s and $e_2$'s. Note that this difference would be
critical if example data were sparse.  We will explain the method used
to compute CCD in Section~\ref{subsec:methodology}.

\subsection{Methodology}
\label{subsec:methodology}

To illustrate the overall algorithm, we will consider an abstract
specification of both an input and the database
(Figure~\ref{fig:case-frame}). Let the input be
\mbox{\{$\Ni{\Ci{1}}$--$\Mi{\Ci{1}}$, $\Ni{\Ci{2}}$--$\Mi{\Ci{2}}$,
$\Ni{\Ci{3}}$--$\Mi{\Ci{3}}$, $\V$\}}, where $\Ni{\Ci{i}}$ denotes the
case filler for the case $\Ci{i}$, and $\Mi{\Ci{i}}$ denotes the case
marker for $\Ci{i}$, and assume that the interpretation candidates for
$\V$ are derived from the database as $\SSi{1}$, $\SSi{2}$ and
$\SSi{3}$. The database also contains a set $\EXi{\SSi{i}}{\Ci{j}}$ of
case filler examples for each case $\Ci{j}$ of each sense $\SSi{i}$
(``---'' indicates that the corresponding case is not allowed).

\begin{figure}
  \small
  \begin{tabular}{|c|ccccl|} \hline
    input & $\Ni{\Ci{1}}$-$\Mi{\Ci{1}}$ & $\Ni{\Ci{2}}$-$\Mi{\Ci{2}}$ &
    $\Ni{\Ci{3}}$-$\Mi{\Ci{3}}$ & & $\V\:(?)$ \\ \hline \hline
    & $\EXi{\SSi{1}}{\Ci{1}}$ &
    $\EXi{\SSi{1}}{\Ci{2}}$ & $\EXi{\SSi{1}}{\Ci{3}}$ &
    --- & $\V\:(\SSi{1}$) \\
    database & $\EXi{\SSi{2}}{\Ci{1}}$ &
    $\EXi{\SSi{2}}{\Ci{2}}$ & $\EXi{\SSi{2}}{\Ci{3}}$ &
    $\EXi{\SSi{2}}{\Ci{4}}$ & $\V\:(\SSi{2}$) \\
    & --- &
    $\EXi{\SSi{3}}{\Ci{2}}$ & $\EXi{\SSi{3}}{\Ci{3}}$ &
    --- & $\V\:(\SSi{3}$) \\ \hline
  \end{tabular}
  \medskip
  \caption{An input and the database.}
  \label{fig:case-frame}
\end{figure}

During the verb sense disambiguation process, the system first
discards those candidates whose case frame does not fit the input. In
the case of Figure~\ref{fig:case-frame}, $\SSi{3}$ is discarded
because the case frame of \mbox{$\V\:(\SSi{3}$)} does not
subcategorize for the case $\Ci{1}$.

In the next step the system computes the score of the remaining
candidates and chooses as the most plausible interpretation the one
with the highest score. The score of an interpretation is computed by
considering the weighted average of the similarity degrees of the
input case fillers with respect to each of the example case fillers
(in the corresponding case) listed in the database for the sense under
evaluation. Formally, this is expressed by Equation~\eq{eq:score},
where \mbox{$Score(\SS)$} is the score of sense $\SS$ of the input
verb, and \mbox{$SIM(\Ni{\C},\EXi{\SS}{\C})$} is the maximum
similarity degree between the input case filler $\Ni{\C}$ and the
corresponding case fillers in the database example set $\EXi{\SS}{\C}$
(calculated through Equation~\eq{eq:sim}). $CCD(\C)$ is the weight
factor of case $\C$, which we will explain later in this section.
\begin{equation}
\label{eq:score}
Score(\SS) = \frac{\textstyle \sum_{\C}
  SIM(\Ni{\C},\EXi{\SS}{\C})\cdot CCD(\C)}{\textstyle \sum_{\C}
  CCD(\C)}
\end{equation}
\begin{equation}
  \label{eq:sim}
  SIM(\Ni{\C},\EXi{\SS}{\C}) = {\displaystyle \max_{\E \in
      \EXi{\SS}{\C}}} sim(\Ni{\C},\E)
\end{equation}

\begin{table}
  \tcaption{The relation between the length of the path between two
  nouns $n_1$ and $n_2$ in the {\it Bunruigoihyo\/} thesaurus
  ($len(n_1,n_2)$), and their relative similarity ($sim(n_1,n_2)$).}
  \label{tab:sim}
  \small
  \begin{tabular}{cccccccc}
    $len(n_1,n_2)$ & 0 & 2 & 4 & 6 & 8 & 10 & 12 \\ \hline
    $sim(n_1,n_2)$ & 11 & 10 & 9 & 8 & 7 & 5 & 0 \\ \hline
  \end{tabular}
\end{table}
  
\begin{figure}
  \centering
  \mbox{\psfig{figure=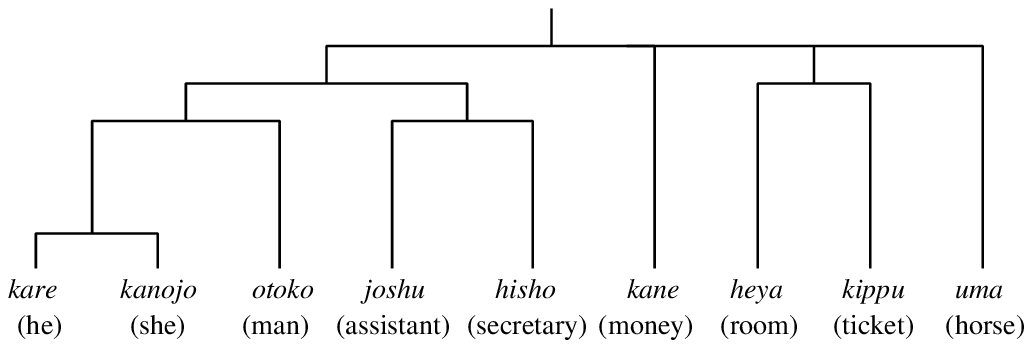,height=1.5in}}
  \medskip
  \caption{A fragment of the {\it Bunruigoihyo\/} thesaurus.}
  \label{fig:bgh}
\end{figure}

With regard to the computation of the similarity between two different
case fillers ($sim(\Ni{\C},\E)$ in Equation~\eq{eq:score}), we
experimentally used two alternative approaches. The first approach
uses semantic resources, that is, hand-crafted thesauri (such as the
Roget's thesaurus~\cite{chapman:84} or
WordNet~\cite{miller:techrep-93} in the case of English, and {\it
Bunruigoihyo\/}~\cite{bgh:64} or EDR~\cite{edr:95} in the case of
Japanese), based on the intuitively feasible assumption that words
located near each other within the structure of a thesaurus have
similar meaning.  Therefore, the similarity between two given words is
represented by the length of the path between them in the thesaurus
structure~\cite{fujii:coling-96,kurohashi:ieice-94,x.li:ijcai-95,uramoto:ieice-94}.\footnote{Different
types of application of hand-crafted thesauri to word sense
disambiguation have been proposed, for example, by
Yarowsky~\shortcite{yarowsky:coling-92}.} We used the similarity
function empirically identified by Kurohashi and Nagao in which the
relation between the length of the path in the {\it Bunruigoihyo\/}
thesaurus and the similarity between words is defined as shown in
Table~\ref{tab:sim}. In this thesaurus, each entry is assigned a
seven-digit class code.  In other words, this thesaurus can be
considered as a tree, seven levels in depth, with each leaf as a set
of words.  Figure~\ref{fig:bgh} shows a fragment of the {\it
Bunruigoihyo\/} thesaurus including some of the nouns in both
Figure~\ref{fig:database} and the input sentence above.

The second approach is based on statistical modeling.  We adopted one
typical implementation called the ``vector space model''
(VSM)~\cite{frakes:92,leacock:arpa-hlt-93,salton:83,schutze:supercomp-92},
which has a long history of application in information retrieval (IR)
and text categorization (TC) tasks.  In the case of IR/TC, VSM is used
to compute the similarity between documents, which is represented by a
vector comprising statistical factors of content words in a
document. Similarly, in our case, each noun is represented by a vector
comprising statistical factors, although statistical factors are
calculated in terms of the predicate-argument structure in which each
noun appears. Predicate-argument structures, which consist of
complements (case filler nouns and case markers) and verbs, have also
been used in the task of noun classification~\cite{hindle:acl-90}.
This can be expressed by Equation~\eq{eq:vector}, where $\vec{n}$ is
the vector for the noun in question, and items $t_{i}$ represent the
statistics for predicate-argument structures including $n$.
\begin{equation}
  \label{eq:vector}
  \vec{n} = <t_{1},~t_{2},~\ldots,~t_{i},~\ldots>
\end{equation}
In regard to $t_{i}$, we used the notion of
TF$\cdot$IDF~\cite{salton:83}. TF (term frequency) gives each context
(a case marker/verb pair) importance proportional to the number of
times it occurs with a given noun. The rationale behind IDF (inverse
document frequency) is that contexts which rarely occur over
collections of nouns are valuable, and that therefore the IDF of a
context is inversely proportional to the number of noun types that
appear in that context.  This notion is expressed by
Equation~\eq{eq:tf_idf}, where \mbox{$f(<\!n,c,v\!>)$} is the
frequency of the tuple \mbox{$<\!n,c,v\!>$}, \mbox{$nf(<\!c,v\!>)$} is
the number of noun types which collocate with verb $v$ in the case
$c$, and $N$ is the number of noun types within the overall
co-occurrence data.
\begin{equation}
  \label{eq:tf_idf}
  t_{i} = f(<\!n,c,v\!>)\cdot\log\frac{\textstyle N}{\textstyle
    nf(<\!c,v\!>)}
\end{equation}
We then compute the similarity between nouns $n_1$ and $n_2$ by the
cosine of the angle between the two vectors $\vec{n_1}$ and
$\vec{n_2}$. This is realized by Equation~\eq{eq:vsm}.
\begin{equation}
  \label{eq:vsm}
  sim(n_1,n_2) = \frac{\textstyle \vec{n_1}\cdot \vec{n_2}}{\textstyle
    |\vec{n_1}||\vec{n_2}|}
\end{equation}
We extracted co-occurrence data from the RWC text base
RWC-DB-TEXT-95-1~\cite{rwc:95}. This text base consists of four years
worth of Mainichi Shimbun newspaper articles~\cite{mainichi:91-94},
which have been automatically annotated with morphological tags. The
total morpheme content is about one hundred million. Since full
parsing is usually expensive, a simple heuristic rule was used to
obtain collocations of nouns, case markers, and verbs in the form of
tuples \mbox{$<\!n,c,v\!>$}. This rule systematically associates each
sequence of noun and case marker to the verb of highest proximity, and
produced 419,132 tuples. This co-occurrence data was used in the
preliminary experiment described in
Section~\ref{subsec:pre_experiment}.\footnote{Note that each verb in
co-occurrence data should ideally be annotated with its verb sense.
However, there is no existing Japanese text base with sufficient
volume of word sense tags.}

In Equation~\eq{eq:score}, \mbox{$CCD(\C)$} expresses the weight
factor of the contribution of case $\C$ to (current) verb sense
disambiguation.  Intuitively, preference should be given to cases
displaying case fillers that are classified in semantic categories of
greater disjunction.  Thus, $\C$'s contribution to the sense
disambiguation of a given verb, \mbox{$CCD(\C)$}, is likely to be
higher if the example case filler sets \mbox{\{$\EXi{\SSi{i}}{\C}~|~i
= 1, \ldots, n$\}} share fewer elements as in Equation~\eq{eq:ccd}.
\begin{equation}
  \begin{array}{l}
    CCD(\C) = {\displaystyle \left(\frac {\displaystyle
        1}{\displaystyle _{\it n\/}{\rm C}_{2}}\sum_{{\it i\/} =
        1}^{{\it n\/}-1}\sum_{{\it j\/} = {\it i\/} + 1}^{{\it
          n\/}}\frac{\displaystyle
        |\EXi{\SSi{i}}{\C}|+|\EXi{\SSi{j}}{\C}|-2|\EXi{\SSi{i}}{\C}\cap
        \EXi{\SSi{j}}{\C}|}{\displaystyle
        |\EXi{\SSi{i}}{\C}|+|\EXi{\SSi{j}}{\C}|}\right)^\alpha
      \label{eq:ccd}}
  \end{array}
\end{equation}
Here, $\alpha$ is a constant for parameterizing the extent to which
CCD influences verb sense disambiguation.  The larger $\alpha$, the
stronger CCD's influence on the system output. To avoid data
sparseness, we smooth each element (noun example) in
$\EXi{\SSi{i}}{\C}$. In practice, this involves generalizing each
example noun into a five-digit class based on the {\it Bunruigoihyo\/}
thesaurus, as has been commonly used for smoothing.

\subsection{Preliminary Experimentation}
\label{subsec:pre_experiment}

We estimated the performance of our verb sense disambiguation method
through an experiment, in which we compared the following five
methods:
\begin{itemize}
\item lower bound (LB), in which the system systematically chooses the
  most frequently appearing verb sense in the
  database~\cite{gale:acl-92},
\item rule-based method (RB), in which the system uses a thesaurus to
  (automatically) identify appropriate semantic classes as selectional
  restrictions for each verb complement,
\item Naive-Bayes method (NB), in which the system interprets a given
  verb based on the probability that it takes each
  verb sense,
\item example-based method using the vector space model (VSM), in
  which the system uses the above mentioned co-occurrence data
  extracted from the RWC text base,
\item example-based method using the {\it Bunruigoihyo\/} thesaurus
  (BGH), in which the system uses Table~\ref{tab:sim} for the
  similarity computation.
\end{itemize}

In the rule-based method, the selectional restrictions are represented
by thesaurus classes, and allow only those nouns dominated by the
given class in the thesaurus structure as verb complements.  In order
to identify appropriate thesaurus classes, we used the association
measure proposed by Resnik~\shortcite{resnik:phd-93}, which computes
the information-theoretic association degree between case fillers and
thesaurus classes, for each verb sense
(Equation~\eq{eq:assoc}).\footnote{Note that previous research has
  applied this technique to tasks other than verb sense
  disambiguation, such as syntactic
  disambiguation~\cite{resnik:phd-93} and disambiguation of case
  filler noun senses~\cite{ribas:eacl-95}.}
\begin{equation}
  \label{eq:assoc}
  A(\SS,\C,r) = P(r|\SS,\C)\cdot\log\frac{\textstyle
    P(r|\SS,\C)}{\textstyle P(r|\C)}
\end{equation}
Here, \mbox{$A(\SS,\C,r)$} is the association degree between verb
sense $\SS$ and class $r$ (selectional restriction candidate) with
respect to case $\C$. \mbox{$P(r|\SS,\C)$} is the conditional
probability that a case filler example associated with case $\C$ of
sense $\SS$ is dominated by class $r$ in the thesaurus.
\mbox{$P(r|\C)$} is the conditional probability that a case filler
example for case $\C$ (disregarding verb sense) is dominated by class
$r$.  Each probability is estimated based on training data.  We used
the semantic classes defined in the {\it Bunruigoihyo\/} thesaurus. In
practice, every $r$ whose association degree is above a certain
threshold is chosen as a selectional
restriction~\cite{resnik:phd-93,ribas:eacl-95}.  By decreasing the
value of the threshold, system coverage can be broadened, but this
opens the way for irrelevant (noisy) selectional rules.

The Naive-Bayes method assumes that each case filler included in a
given input is conditionally independent of other case fillers: the
system approximates the probability that an input $x$ takes a verb
sense $\SS$ (\mbox{$P(\SS|x)$}), simply by computing the product of
the probability that each verb sense $\SS$ takes $\Ni{\C}$ as a case
filler for case $\C$. The verb sense with maximal probability is then
selected as the interpretation (Equation~\eq{eq:prob}).\footnote{A
number of experimental results have shown the effectiveness of the
Naive-Bayes method for word sense
disambiguation~\cite{gale:ch-92,leacock:arpa-hlt-93,mooney:emnlp-96,ng:emnlp-97,pedersen:anlp-97}.}
\begin{equation}
  \label{eq:prob}
  \begin{array}{lll}
    {\displaystyle \arg\max_{\SS} P(\SS|x)} & = & {\displaystyle
      \arg\max_{\SS} \frac{\textstyle P(\SS)\cdot P(x|\SS)}{\textstyle
        P(x)}} \\ \noalign{\vskip 1ex} & = & {\displaystyle
      \arg\max_{\SS} P(\SS)\cdot P(x|\SS)} \\ \noalign{\vskip 1ex} &
    \approx & {\displaystyle \arg\max_{\SS} P(\SS) \prod_{\C}
      P(\Ni{\C}|\SS)}
  \end{array}
\end{equation}
Here, \mbox{$P(\Ni{\C}|\SS)$} is the probability that a case filler
associated with sense $\SS$ for case $\C$ in the training data is
$\Ni{\C}$. We estimated $P(\SS)$ based on the distribution of the verb
senses in the training data.  In practice, data sparseness leads to
not all case fillers $\Ni{\C}$ appearing in the database, and as such,
we generalize each $\Ni{\C}$ into semantic class defined in the {\it
Bunruigoihyo\/} thesaurus.

All methods excepting the lower bound method involve a parametric
constant: the threshold value for the association degree (RB), a
generalization level for case filler nouns (NB), and $\alpha$ in
Equation~\eq{eq:ccd} (VSM and BGH). For these parameters, we conducted
several trials prior to the actual comparative experiment, to
determine the optimal parameter values over a range of data sets.  For
our method, we set $\alpha$ extremely large, which is equivalent to
relying almost solely on the SIM of the case with greatest
CCD. However, note that when the SIM of the case with greatest CCD is
equal for multiple verb senses, the system computes the SIM of the
case with second highest CCD. This process is repeated until only one
verb sense remains.  When more than one verb sense is selected for any
given method (or none of them remains, for the rule-based method), the
system simply selects the verb sense that appears most frequently in
the database.\footnote{One may argue that this goes against the basis
of the rule-based method, in that, given a proper threshold value for
the association degree, the system could improve on accuracy
(potentially sacrificing coverage), and that the trade-off between
coverage and accuracy is therefore a more appropriate evaluation
criterion.  However, our trials on the rule-based method with
different threshold values did not show significant correlation
between the improvement of accuracy and the degeneration of the
coverage.}

In the experiment, we conducted sixfold cross-validation, that is, we
divided the training/test data into six equal parts, and conducted six
trials in which a different part was used as test data each time, and
the rest as training data (the database).\footnote{Ideally speaking,
training and test data should be drawn from different sources, to
simulate a {\it real\/} application. However, the sentences were
already scrambled when provided to us, and therefore we could not
identify the original source corresponding to each sentence.} We
evaluated the performance of each method according to its accuracy,
that is the ratio of the number of correct outputs compared to the
total number of inputs.  The training/test data used in the experiment
contained about one thousand simple Japanese sentences collected from
news articles. Each sentence in the training/test data contained one
or more complement(s) followed by one of the eleven verbs described in
Table~\ref{tab:corpus}. In Table~\ref{tab:corpus}, the column of
``English Gloss'' describes typical English translations of the
Japanese verbs.  The column of ``\# of Sentences'' denotes the number
of sentences in the corpus, and ``\# of Senses'' denotes the number of
verb senses contained in IPAL. The column of ``Accuracy'' shows the
accuracy of each method.

\begin{table}
  \tcaption{The verbs contained in the corpus used, and the accuracy of
    the different verb sense disambiguation methods (LB: lower bound,
    RB: rule-based method, NB: Naive-Bayes method, VSM: vector space
    model, BGH: the {\it Bunruigoihyo\/} thesaurus).}
  \small
  \begin{tabular}{ccccccccc}
    & & \# of & \# of & \multicolumn{5}{c}{Accuracy (\%)}
    \\ \cline{5-9}
    Verb & English Gloss & Sentences & Senses & LB & RB & NB & VSM &
    BGH \\ \hline \hline
    {\it ataeru\/} & give & 136 & 4 &
    ~66.9~ & ~62.1~ & ~75.8~ & ~84.1~ & ~86.0~ \\ \hline
    {\it kakeru\/} & hang & 160 & 29 &
    25.6 & 24.6 & 67.6 & 73.4 & 76.2 \\ \hline
    {\it kuwaeru\/} & add & 167 & 5 &
    53.9 & 65.6 & 82.2 & 84.0 & 86.8 \\ \hline
    {\it motomeru\/} & require & 204 & 4 &
    85.3 & 82.4 & 87.0 & 85.5 & 85.5 \\ \hline
    {\it noru\/} & ride & 126 & 10 &
    45.2 & 52.8 & 81.4 & 80.5 & 85.3 \\ \hline
    {\it osameru\/} & govern & 108 & 8 &
    30.6 & 45.6 & 66.0 & 72.0 & 74.5 \\ \hline
    {\it tsukuru\/} & make & 126 & 15 &
    25.4 & 24.9 & 59.1 & 56.5 & 69.9 \\ \hline
    {\it toru\/} & take & 84 & 29 &
    26.2 & 16.2 & 56.1 & 71.2 & 75.9 \\ \hline
    {\it umu\/} & bear offspring & 90 & 2 &
    83.3 & 94.7 & 95.5 & 92.0 & 99.4 \\ \hline
    {\it wakaru\/} & understand & 60 & 5 &
    48.3 & 40.6 & 71.4 & 62.5 & 70.7 \\ \hline
    {\it yameru\/} & stop & 54 & 2 &
    59.3 & 89.9 & 92.3 & 96.2 & 96.3 \\ \hline \hline
    total & --- & 1,315 & --- &
    51.4 & 54.8 & 76.6 & 78.6 & 82.3 \\ \hline
  \end{tabular}
  \label{tab:corpus}
\end{table}

Looking at Table~\ref{tab:corpus}, one can see that our example-based
method performed better than the other methods (irrespective of the
similarity computation), although the Naive-Bayes method is relatively
comparable in performance. Surprisingly, despite the relatively ad hoc
similarity definition utilized (see Table~\ref{tab:sim}), the {\it
Bunruigoihyo\/} thesaurus led to a greater accuracy gain than the
vector space model.  In order to estimate the upper bound (limitation)
of the disambiguation task, that is, to what extent a human expert
makes errors in disambiguation~\cite{gale:acl-92}, we analyzed
incorrect outputs and found that roughly 30\% of the system errors
using the {\it Bunruigoihyo\/} thesaurus fell into this category.  It
should be noted that while the vector space model requires
computational cost (time/memory) of an order proportional to the size
of the vector, determination of paths in the {\it Bunruigoihyo\/}
thesaurus comprises a trivial cost.

We also investigated errors made by the rule-based method to find a
rational explanation for its inferiority.  We found that the
association measure in Equation~\eq{eq:assoc} tends to give a greater
value to less frequently appearing verb senses and lower level (more
specified) classes, and therefore chosen rules are generally
overspecified.\footnote{This problem has also been identified by
Charniak~\shortcite{charniak:93}.} Consequently, frequently appearing
verb senses are likely to be rejected. On the other hand, when
attempting to enhance the rule set by setting a smaller threshold
value for the association score, overgeneralization can be a problem.
We also note that one of the theoretical differences between the
rule-based and example-based methods is that the former statically
generalizes examples (prior to system usage), while the latter does so
dynamically. Static generalization would appear to be relatively risky
for sparse training data.

Although comparison of different approaches to word sense
disambiguation should be further investigated, this experimental
result gives us good motivation to explore example-based verb sense
disambiguation approaches, i.e., to introduce the notion of selective
sampling into them.

\subsection{Enhancement of Verb Sense Disambiguation}
\label{subsec:enhancement}

Let us discuss how further enhancements to our example-based verb
sense disambiguation system could be made. First, since inputs are
simple sentences, information for word sense disambiguation is
inadequate in some cases. External information such as the discourse
or domain dependency of each word
sense~\cite{guthrie:acl-91,nasukawa:tmi-93,yarowsky:acl-95} is
expected to lead to system improvement.  Second, some idiomatic
expressions represent highly restricted collocations, and
overgeneralizing them semantically through the use of a thesaurus can
cause further errors. Possible solutions would include one proposed by
Uramoto, in which idiomatic expressions are described separately in
the database so that the system can control their
overgeneralization~\cite{uramoto:ieice-94}.  Third, a number of
existing NLP tools such as the JUMAN (morphological
analyzer)~\cite{matsumoto:93} and QJP (morphological and syntactic
analyzer)~\cite{kameda:coling-96} could broaden the coverage of our
system, as inputs are currently limited to simple, morphologically
analyzed sentences. Finally, it should be noted that in Japanese, case
markers can be omitted or topicalized (for example, marked with
postposition {\it wa\/}), an issue which our framework does not
currently consider.

\section{Example Sampling Algorithm}
\label{sec:sampling}

\subsection{Overview}
\label{subsec:overview}

Let us look again at Figure~\ref{fig:concept} in
Section~\ref{sec:intro}. In this figure, ``WSD outputs'' refers to a
corpus in which each sentence is assigned an expected verb
interpretation during the WSD phase. In the training phase, the system
stores supervised samples (with each interpretation simply checked or
appropriately corrected by a human) in the database, to be used in a
later WSD phase. In this section, we turn to the problem as to which
examples should be selected as samples.

Lewis and Gale~\shortcite{lewis:sigir-94} proposed the notion of
uncertainty sampling for the training of statistics-based text
classifiers. Their method selects those examples that the system
classifies with minimum certainty, based on the assumption that there
is no need for teaching the system the correct answer when it has
answered with sufficiently high certainty. However, we should take
into account the training effect a given example has on other
remaining (unsupervised) examples. In other words, we would like to
select samples such as to be able to correctly disambiguate as many
examples as possible in the next iteration.  If this is successfully
done, the number of examples to be supervised will decrease. We
consider maximization of this effect by means of a training utility
function aimed at ensuring that the most useful example at a given
point in time is the example with the greatest training utility
factor. Intuitively speaking, the training utility of an example is
greater when we can expect greater increase in the interpretation
certainty of the remaining examples after training using that example.

To explain this notion intuitively, let us take
Figure~\ref{fig:yameru} as an example corpus.  In this corpus, all
sentences contain the verb {\it yameru}, which has two senses
according to IPAL, $s_1$ (`to stop (something)') and $s_2$ (`to quit
(occupation)'). In this figure, sentences $e_1$ and $e_2$ are
supervised examples associated with the senses $s_1$ and $s_2$,
respectively, and $x_i$'s are unsupervised examples.  For the sake of
enhanced readability, the examples $x_i$'s are partitioned according
to their verb senses, that is, $x_1$ to $x_5$ correspond to sense
$s_1$, and $x_6$ to $x_9$ correspond to sense $s_2$.  In addition,
note that examples in the corpus can be readily categorized based on
case similarity, that is, into clusters \mbox{$\{x_1,x_2,x_3,x_4\}$}
(`someone/something stops service'), \mbox{$\{e_2,x_6,x_7\}$}
(`someone leaves organization'), \mbox{$\{x_8,x_9\}$} (`someone quits
occupation'), $\{e_1\}$, and $\{x_5\}$.  Let us simulate the sampling
procedure with this example corpus.  In the initial stage with
\mbox{$\{e_1,e_2\}$} in the database, $x_6$ and $x_7$ can be
interpreted as $s_2$ with greater certainty than for other $x_i$'s,
because these two examples are similar to $e_2$.  Therefore,
uncertainty sampling selects any example excepting $x_6$ and $x_7$ as
the sample.  However, any one of examples $x_1$ to $x_4$ is more
desirable because by way of incorporating one of these examples, we
can obtain more $x_i$'s with greater certainty. Assuming that $x_1$ is
selected as the sample and incorporated into the database with sense
$s_1$, either of $x_8$ and $x_9$ will be more highly desirable than
other unsupervised $x_i$'s in the next stage.

\begin{figure}
  \small
  \begin{tabular}{|clll|} \hline
    $e_1$: & {\it seito ga\/} (student-NOM) & {\it shitsumon wo\/}
    (question-ACC) & {\it yameru\/} ($\SSi{1}$) \\
    $e_2$: & {\it ani ga\/} (brother-NOM) & {\it kaisha~wo\/}
    (company-ACC) & {\it yameru\/} ($\SSi{2}$) \\ \hline \hline
    $x_1$: & {\it shain ga\/} (employee-NOM) & {\it eigyou wo\/}
    (sales-ACC) & {\it yameru\/} (?) \\
    $x_2$: & {\it shouten ga\/} (store-NOM) & {\it eigyou wo\/}
    (sales-ACC) & {\it yameru\/} (?) \\
    $x_3$: & {\it koujou ga\/} (factory-NOM) & {\it sougyou wo\/}
    (operation-ACC) & {\it yameru\/} (?) \\
    $x_4$: & {\it shisetsu ga\/} (facility-NOM) & {\it unten wo\/}
    (operation-ACC) & {\it yameru\/} (?) \\
    $x_5$: & {\it sensyu ga\/} (athlete-NOM) & {\it renshuu wo\/}
    (practice-ACC) & {\it yameru\/} (?) \\
    $x_6$: & {\it musuko ga\/} (son-NOM) & {\it kaisha wo\/}
    (company-ACC) & {\it yameru\/} (?) \\
    $x_7$: & {\it kangofu ga\/} (nurse-NOM) & {\it byouin wo\/}
    (hospital-ACC) & {\it yameru\/} (?) \\
    $x_8$: & {\it hikoku ga\/} (defendant-NOM) & {\it giin wo\/}
    (congressman-ACC) & {\it yameru\/} (?) \\
    $x_9$: & {\it chichi ga\/} (father-NOM) & {\it kyoushi wo\/}
    (teacher-ACC) & {\it yameru\/} (?) \\ \hline
  \end{tabular}
  \medskip
  \caption{Example of a given corpus associated with the verb {\it
      yameru}.}
  \label{fig:yameru}
\end{figure}

Let $\set{S}$ be a set of sentences, i.e., a given corpus, and
$\set{D}$ be the subset of supervised examples stored in the database.
Further, let $\set{X}$ be the set of unsupervised examples, realizing
Equation~\eq{eq:corpus}.
\begin{equation}
  \label{eq:corpus}
  \set{S} = \set{D} \cup \set{X}
\end{equation}
The example sampling procedure can be illustrated as:
\begin{enumerate}
\item $WSD(\set{D}, \set{X})$
\item $e \leftarrow \arg\max_{x\in\set{X}}TU(x)$ 
\item $\set{D} \leftarrow \set{D} \cup\{e\},~~\set{X}
  \leftarrow \set{X}\cap \overline{\{e\}}$
\item goto 1
\end{enumerate}
where \mbox{$WSD(\set{D}, \set{X})$} is the verb sense disambiguation
process on input $\set{X}$ using $\set{D}$ as the database. In this
disambiguation process, the system outputs the following for each
input: (a) a set of verb sense candidates with interpretation scores,
and (b) an interpretation certainty. These factors are used for the
computation of \mbox{$TU(x)$}, newly introduced in our method. 
\mbox{$TU(x)$} computes the training utility factor for an example
$x$.  The sampling algorithm gives preference to examples of maximum
utility.

We will explain in the following sections how \mbox{$TU(x)$} is
estimated, based on the estimation of the interpretation certainty.

\subsection{Interpretation Certainty}
\label{subsec:certainty}

Lewis and Gale~\shortcite{lewis:sigir-94} estimate certainty of an
interpretation as the ratio between the probability of the most
plausible text category and the probability of any other text
category, excluding the most probable one.  Similarly, in our verb
sense disambiguation system, we introduce the notion of interpretation
certainty of examples based on the following preference conditions:
\begin{enumerate}
\item the highest interpretation score is greater,
\item the difference between the highest and second highest
  interpretation scores is greater.
\end{enumerate}
The rationale for these conditions is given below. Consider
Figure~\ref{fig:certainty}, where each symbol denotes an example in a
given corpus, with symbols $x$ as unsupervised examples and symbols
$e$ as supervised examples.  The curved lines delimit the semantic
vicinities (extents) of the two verb senses 1 and 2,
respectively.\footnote{Note that this method can easily be extended
for a verb which has more than two senses. In Section~\ref{sec:eval},
we describe an experiment using multiply polysemous verbs.} The
semantic similarity between two examples is graphically portrayed by
the physical distance between the two symbols representing them. In
Figure~\ref{fig:certainty}(a), $x$'s located inside a semantic
vicinity are expected to be interpreted as being similar to the
appropriate example $e$ with high certainty, a fact which is in line
with condition~1 above. However, in Figure~\ref{fig:certainty}(b), the
degree of certainty for the interpretation of any $x$ located inside
the intersection of the two semantic vicinities cannot be great.  This
occurs when the case fillers associated with two or more verb senses
are not selective enough to allow for a clear-cut delineation between
them. This situation is explicitly rejected by condition~2.

\begin{figure}
  \mbox{
    \begin{minipage}[t]{.47\textwidth}
      \centering
      \psfig{figure=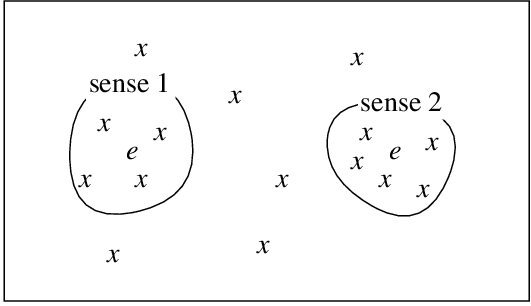,height=1.5in}
      (a)
    \end{minipage}
    \hfill
    \begin{minipage}[t]{.47\textwidth}
      \centering
      \psfig{figure=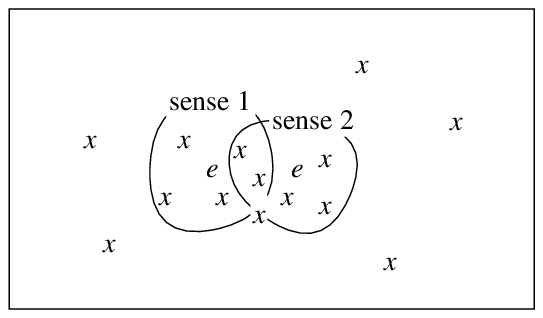,height=1.5in}
      (b)
    \end{minipage}}
  \medskip
  \caption{The concept of interpretation certainty. The case where the
  interpretation certainty of the enclosed $x$'s is great is shown in
  (a). The case where the interpretation certainty of the $x$'s
  contained in the intersection of senses 1 and 2 is small is shown in
  (b).}
  \label{fig:certainty}
\end{figure}

Based on the above two conditions, we compute interpretation
certainties using Equation~\eq{eq:certainty}, where \mbox{$C(x)$} is
the interpretation certainty of an example $x$. \mbox{$Score_1(x)$}
and \mbox{$Score_2(x)$} are the highest and second highest scores for
$x$, respectively, and $\lambda$, which ranges from 0 to 1, is a
parametric constant used to control the degree to which each condition
affects the computation of \mbox{$C(x)$}.
\begin{equation}
  \label{eq:certainty}
  C(x) = \lambda\cdot Score_1(x) + (1 - \lambda)\cdot(Score_1(x) -
  Score_2(x))
\end{equation}

Through a preliminary experiment, we estimated the validity of the
notion of the interpretation certainty, by the trade-off between
accuracy and coverage of the system. Note that in this experiment,
accuracy is the ratio of the number of correct outputs and the number
of cases where the interpretation certainty of the output is above a
certain threshold. Coverage is the ratio of the number of cases where
the interpretation certainty of the output is above a certain
threshold and the number of inputs.  By raising the value of the
threshold, accuracy also increases (at least theoretically), while
coverage decreases.

\begin{figure}
  \centering
  \mbox{\psfig{figure=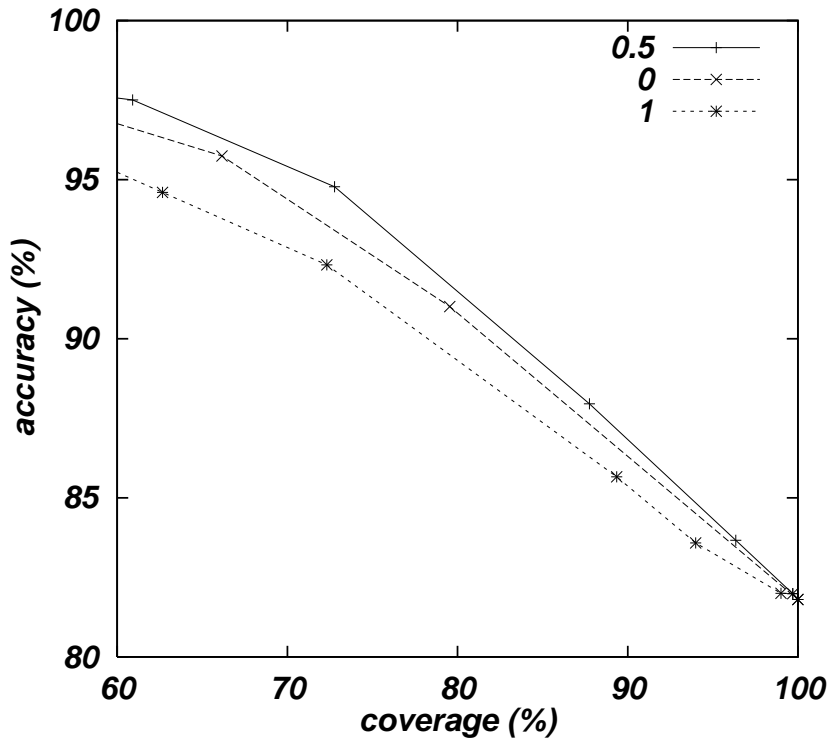,height=4in}}
  \medskip
  \caption{The relation between coverage and accuracy with different
    $\lambda$'s.}
  \label{fig:app-acc}
\end{figure}

The system used the {\it Bunruigoihyo\/} thesaurus for the similarity
computation, and was evaluated by way of sixfold cross-validation
using the same corpus as that used for the experiment described in
Section~\ref{subsec:pre_experiment}.  Figure~\ref{fig:app-acc} shows
the result of the experiment with several values of $\lambda$, from
which the optimal $\lambda$ value seems to be in the range around 0.5. 
It can be seen that, as we assumed, both of the above conditions are
essential for the estimation of the interpretation certainty.

\begin{figure}
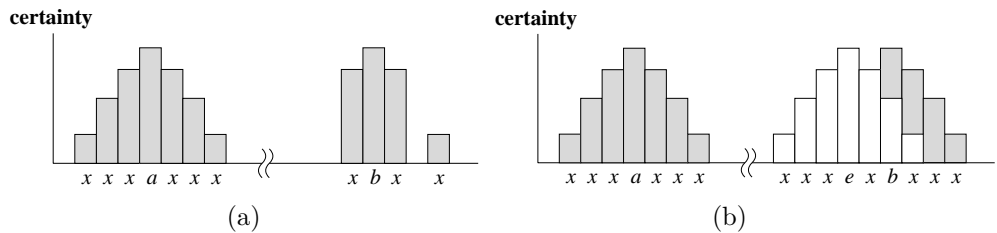

  \mbox{
    \begin{minipage}[t]{.47\textwidth}
      \centering
      \psfig{figure=tu-a.eps,height=1in}
      (a)
    \end{minipage}
    \hfill
    \begin{minipage}[t]{.47\textwidth}
      \centering
      \psfig{figure=tu-b.eps,height=1in}
      (b)
    \end{minipage}}
  \medskip
  \caption{The concept of training utility. The case where the
  training utility of $a$ is greater than that of $b$ because $a$ has
  more unsupervised neighbors is shown in (a); (b) shows the case
  where the training utility of $a$ is greater than that of $b$
  because $b$ closely neighbors $e$, contained in the database.}
  \label{fig:tu}
\end{figure}

\subsection{Training Utility}
\label{subsec:utility}

The training utility of an example $a$ is greater than that of another
example $b$ when the total interpretation certainty of unsupervised
examples increases more after training with example $a$ than with
example $b$. Let us consider Figure~\ref{fig:tu}, in which the x-axis
mono-dimensionally denotes the semantic similarity between two
unsupervised examples, and the y-axis denotes the interpretation
certainty of each example.  Let us compare the training utility of the
examples $a$ and $b$ in Figure~\ref{fig:tu}(a).  Note that in this
figure, whichever example we use for training, the interpretation
certainty for each unsupervised example ($x$) neighboring the chosen
example increases based on its similarity to the supervised
example. Since the increase in the interpretation certainty of a given
$x$ becomes smaller as the similarity to $a$ or $b$ diminishes, the
training utility of the two examples can be represented by the shaded
areas. The training utility of $a$ is greater as it has more neighbors
than $b$.  On the other hand, in Figure~\ref{fig:tu}(b), $b$ has more
neighbors than $a$.  However, since $b$ is semantically similar to
$e$, which is already contained in the database, the total {\it
increase\/} in interpretation certainty of its neighbors, i.e., the
training utility of $b$, is smaller than that of $a$.

Let \mbox{$\Delta C(x\!=\!\SS,y)$} be the difference in the
interpretation certainty of \mbox{$y\in\set{X}$} after training with
\mbox{$x\in\set{X}$}, taken with the sense
$\SS$. \mbox{$TU(x\!=\!\SS)$}, which is the training utility function
for $x$ taken with sense $\SS$, can be computed by way of
Equation~\eq{eq:utility}.
\begin{equation}
  \label{eq:utility}
  TU(x\!=\!\SS) = \sum_{y \in \set{X}}\Delta C(x\!=\!\SS,y)
\end{equation}
It should be noted that in Equation~\eq{eq:utility}, we can replace
$\set{X}$ with a subset of $\set{X}$ which consists of neighbors of
$x$. However, in order to facilitate this, an efficient algorithm to
search for neighbors of an example is required. We will discuss this
problem in Section~\ref{subsec:discussion}.

Since there is no guarantee that $x$ will be supervised with any given
sense $\SS$, it can be risky to rely solely on \mbox{$TU(x\!=\!\SS)$}
for the computation of \mbox{$TU(x)$}. We estimate \mbox{$TU(x)$} by
the expected value of $x$, calculating the average of each
\mbox{$TU(x\!=\!\SS)$}, weighted by the probability that $x$ takes
sense $\SS$. This can be realized by Equation~\eq{eq:utility_sum},
where \mbox{$P(\SS|x)$} is the probability that $x$ takes the sense
$\SS$.
\begin{equation}
  \label{eq:utility_sum}
  TU(x) = \sum_{\SS}P(\SS|x)\cdot TU(x\!=\!\SS)
\end{equation}
Given the fact that (a) \mbox{$P(\SS|x)$} is difficult to estimate in
the current formulation, and (b) the cost of computation for each
\mbox{$TU(x\!=\!\SS)$} is not trivial, we temporarily approximate
\mbox{$TU(x)$} as in Equation~\eq{eq:utility_temp}, where $\set{K}$ is
a set of the $k$-best verb sense(s) of $x$ with respect to the
interpretation score in the current state.
\begin{equation}
  \label{eq:utility_temp}
  TU(x) \approx \frac{\textstyle 1}{\textstyle k}\sum_{\SS \in \set{K}}
  TU(x\!=\!\SS)
\end{equation}

\subsection{Enhancement of computation}
\label{subsec:computation}

In this section, we discuss how to enhance the computation associated
with our example sampling algorithm.

\begin{figure}
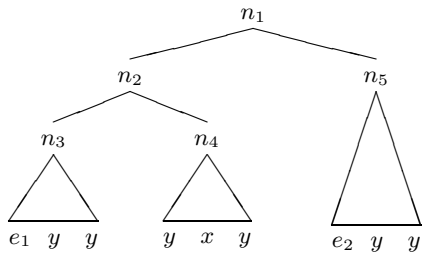

  \small
  \centering
  \unitlength=.05ex
  \tree{
    \node{$n_1$}
    {\Ln7{$n_2$}
      {\Ln5{$n_3$}\tangle8{$e_1$~~$y$~~~$y$}}
      {\Rn5{$n_4$}\tangle8{$y$~~~$x$~~~$y$}}}
    {\Rn7{$n_5$}
      \tangle4{$e_2$~~$y$~~~$y$}}}
  \medskip
  \caption{A fragment of the thesaurus including neighbors of $x$
    associated with case $\C$.}
  \label{fig:neighbor}
\end{figure}

First, we note that computation of \mbox{$TU(x\!=\!\SS)$} in
Equation~\eq{eq:utility} above becomes time consuming because the
system is required to search the whole set of unsupervised examples
for examples whose interpretation certainty will increase after $x$ is
used for training.  To avoid this problem, we could potentially apply
a method used in efficient database search techniques, by which the
system can search for neighbor examples of $x$ with optimal time
complexity~\cite{utsuro:coling-94}.  However, in this section, we will
explain another efficient algorithm to identify neighbors of $x$, in
which neighbors of case fillers are considered to be given directly by
the thesaurus structure.\footnote{Utsuro's method requires the
construction of large-scale similarity templates prior to similarity
computation~\cite{utsuro:coling-94}, and this is what we would like to
avoid.} The basic idea is the following: the system searches for
neighbors of each case filler of $x$ instead of $x$ as a whole, and
merges them as a set of neighbors of $x$.  Note that by dividing
examples along the lines of each case filler, we can retrieve
neighbors based on the structure of the {\it Bunruigoihyo\/} thesaurus
(instead of the conceptual semantic space as in
Figure~\ref{fig:certainty}).  Let $\seti{N}{x=\SS,\C}$ be a subset of
unsupervised neighbors of $x$ whose interpretation certainty will
increase after $x$ is used for training, considering only case $\C$ of
sense $\SS$. The {\it actual\/} neighbor set of $x$ with sense $\SS$
($\seti{N}{x=\SS}$) is then defined as in Equation~\eq{eq:neighbor}.
\begin{equation}
  \label{eq:neighbor}
  \seti{N}{x=\SS} = \bigcup_{\C}\seti{N}{x=\SS,\C}
\end{equation}
Figure~\ref{fig:neighbor} shows a fragment of the thesaurus, in which
$x$ and the $y$'s are unsupervised case filler examples.  Symbols
$e_1$ and $e_2$ are case filler examples stored in the database taken
as senses $\SSi{1}$ and $\SSi{2}$, respectively. The triangles
represent subtrees of the structure, and the labels $n_i$ represent
nodes.  In this figure, it can easily be seen that the interpretation
score of $\SSi{1}$ never changes for examples other than the children
of $n_4$, after $x$ is used for training with sense $\SSi{1}$. In
addition, incorporating $x$ into the database with sense $\SSi{1}$
never changes the score of examples $y$ for other sense candidates.
Therefore, $\seti{N}{x=\SSi{1},\C}$ includes only examples dominated
by $n_4$, in other words, examples that are more closer to $x$ than
$e_1$ in the thesaurus structure.  Since, during the WSD phase, the
system determines $e_1$ as the supervised neighbor of $x$ for sense
$\SSi{1}$, identifying $\seti{N}{x=\SSi{1},\C}$ does not require any
extra computational overhead. We should point out that the technique
presented here is not applicable when the vector space model (see
Section~\ref{subsec:methodology}) is used for the similarity
computation. However, automatic clustering algorithms, which assign a
hierarchy to a set of words based on the similarity between them (for
example the one proposed by Tokunaga, Iwayama, and
Tanaka~\shortcite{tokunaga:ijcai-95}), could potentially facilitate
the application of this retrieval method to the vector space model.

Second, sample size at each iteration should ideally be one, so as to
avoid the supervision of similar examples. On the other hand, a small
sampling size generates a considerable computation overhead for each
iteration of the sampling procedure.  This can be a critical problem
for statistics-based approaches, as the reconstruction of statistic
classifiers is expensive. However, example-based systems fortunately
do not require reconstruction, and examples simply have to be stored
in the database.  Furthermore, in each disambiguation phase, our
example-based system needs only compute the similarity between each
newly stored example and its unsupervised neighbors, rather than
between every example in the database and every unsupervised example.
Let us reconsider Figure~\ref{fig:neighbor}. As mentioned above, when
$x$ is stored in the database with sense $\SSi{1}$, only the
interpretation score of $y$'s dominated by $n_4$,
i.e., $\seti{N}{x=\SSi{1},\C}$, will be changed with respect to sense
$\SSi{1}$.  This algorithm reduces the time complexity of each
iteration from \mbox{$O(N^2)$} to \mbox{$O(N)$}, given that $N$ is the
total number of examples in a given corpus.

\subsection{Discussion}
\label{subsec:discussion}

\subsubsection{Sense Ambiguity of Case Fillers in Selective Sampling}
\label{subsubsec:nsa}

\begin{figure}
  \mbox{
    \begin{minipage}[t]{.47\textwidth}
      \centering
      \psfig{figure=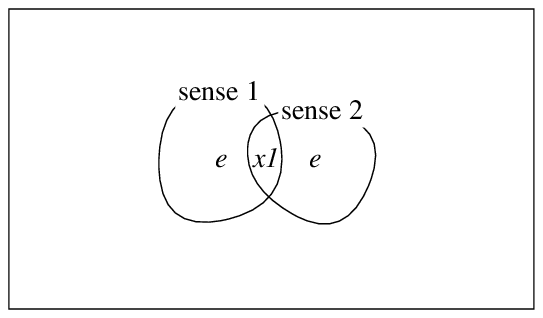,height=1.5in}
      (a)
    \end{minipage}
    \hfill
    \begin{minipage}[t]{.47\textwidth}
      \centering
      \psfig{figure=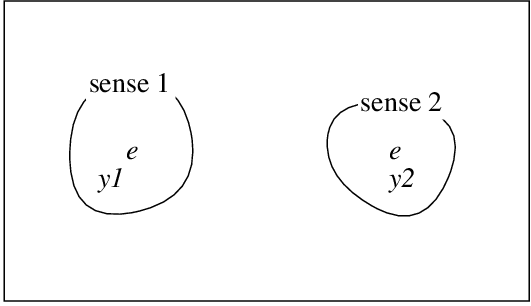,height=1.5in}
      (b)
    \end{minipage}}
  \medskip
  \caption{Two separate scenarios in which the interpretation
  certainty of $x$ is small. In (a), interpretation certainty of $x$
  is small because $x$ lies in the intersection of distinct verb
  senses; in (b), interpretation certainty of $y$ is small because $y$
  is semantically ambiguous.}
  \label{fig:nsa}
\end{figure}

The semantic ambiguity of case fillers (nouns) should be taken into
account during selective sampling. Figure~\ref{fig:nsa}, which uses
the same basic notation as Figure~\ref{fig:certainty}, illustrates one
possible problem caused by case filler ambiguity.  Let $x_1$ be a
sense of a case filler $x$, and $y_1$ and $y_2$ be different senses of
a case filler $y$.  On the basis of Equation~\eq{eq:certainty}, the
interpretation certainty of $x$ and $y$ is small in
Figures~\ref{fig:nsa}(a) and \ref{fig:nsa}(b), respectively.  However,
in the situation shown in Figure~\ref{fig:nsa}(b), since (a) the task
of distinguishing between the {\it verb\/} senses 1 and 2 is easier,
and (b) instances where the sense ambiguity of case fillers
corresponds to distinct verb senses will be rare, training using
either $y_1$ or $y_2$ will be less effective than using a case filler
of the type of $x$.  It should also be noted that since {\it
Bunruigoihyo\/} is a relatively small-sized thesaurus with limited
word sense coverage, this problem is not critical in our case.
However, given other existing thesauri like the EDR electronic
dictionary~\cite{edr:95} or WordNet~\cite{miller:techrep-93}, these
two situations should be strictly differentiated.

\begin{figure}
  \centering
  \mbox{\psfig{figure=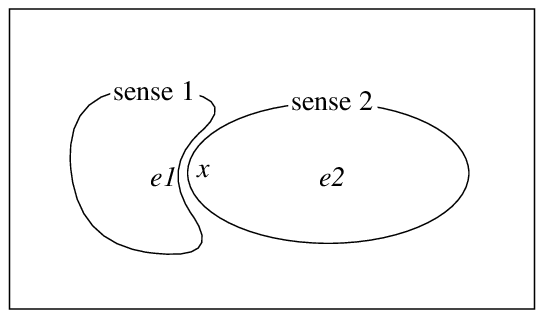,height=1.5in}}
  \medskip
  \caption{The case where informative example $x$ is not selected.}
  \label{fig:limitation}
\end{figure}

\subsubsection{A Limitation of our Selective Sampling Method}
\label{subsubsec:limitation}

Figure~\ref{fig:limitation}, where the basic notation is the same as
in Figure~\ref{fig:certainty}, exemplifies a limitation of our
sampling method. In this figure, the only supervised examples
contained in the database are $e_1$ and $e_2$, and $x$ represents an
unsupervised example belonging to sense 2. Given this scenario, $x$ is
informative because (a) it clearly evidences the semantic vicinity of
sense 2, and (b) without $x$ as sense 2 in the database, the system
may misinterpret other examples neighboring $x$. However, in our
current implementation, the training utility of $x$ would be small
because it would be mistakenly interpreted as sense 1 with great
certainty due to its relatively close semantic proximity to $e_1$.
Even if $x$ has a number of unsupervised neighbors, the total
increment of their interpretation certainty cannot be expected to be
large.  This shortcoming often presents itself when the semantic
vicinities of different verb senses are closely aligned or their
semantic ranges are not disjunctive.  Here, let us consider
Figure~\ref{fig:ccd} again, in which the nominative case would
parallel the semantic space shown in Figure~\ref{fig:limitation} more
closely than the accusative.  Relying more on the similarity in the
accusative (the case with greater CCD) as is done in our system, we
aim to map the semantic space in such a way as to achieve higher
semantic disparity and minimize this shortcoming.

\section{Evaluation}
\label{sec:eval}

\subsection{Comparative Experimentation}
\label{subsec:experiment}

In order to investigate the effectiveness of our example sampling
method, we conducted an experiment, in which we compared the following
four sampling methods:
\begin{itemize}
\item a control (random), in which a certain proportion of a given
  corpus is randomly selected for training,
\item uncertainty sampling (US), in which examples with minimum
  interpretation certainty are selected~\cite{lewis:sigir-94},
\item committee-based sampling (CBS)~\cite{engelson:acl-96},
\item our method based on the notion of training utility (TU).
\end{itemize}
We elaborate on uncertainty sampling and committee-based sampling in
Section~\ref{subsec:related}. We compared these sampling methods by
evaluating the relation between the number of training examples
sampled and the performance of the system.  We conducted sixfold
cross-validation and carried out sampling on the training set. With
regard to the training/test data set, we used the same corpus as that
used for the experiment described in
Section~\ref{subsec:pre_experiment}. Each sampling method uses
examples from IPAL to initialize the system, with the number of
example case fillers for each case being an average of about 3.7.  For
each sampling method, the system uses the {\it Bunruigoihyo\/}
thesaurus for the similarity computation.  In Table~\ref{tab:corpus}
(in Section~\ref{subsec:pre_experiment}), the column of ``accuracy''
for ``BGH'' denotes the accuracy of the system with the entire set of
training data contained in the database. Each of the four sampling
methods achieved this figure at the conclusion of training.

\begin{figure}
  \centering
  \mbox{\psfig{figure=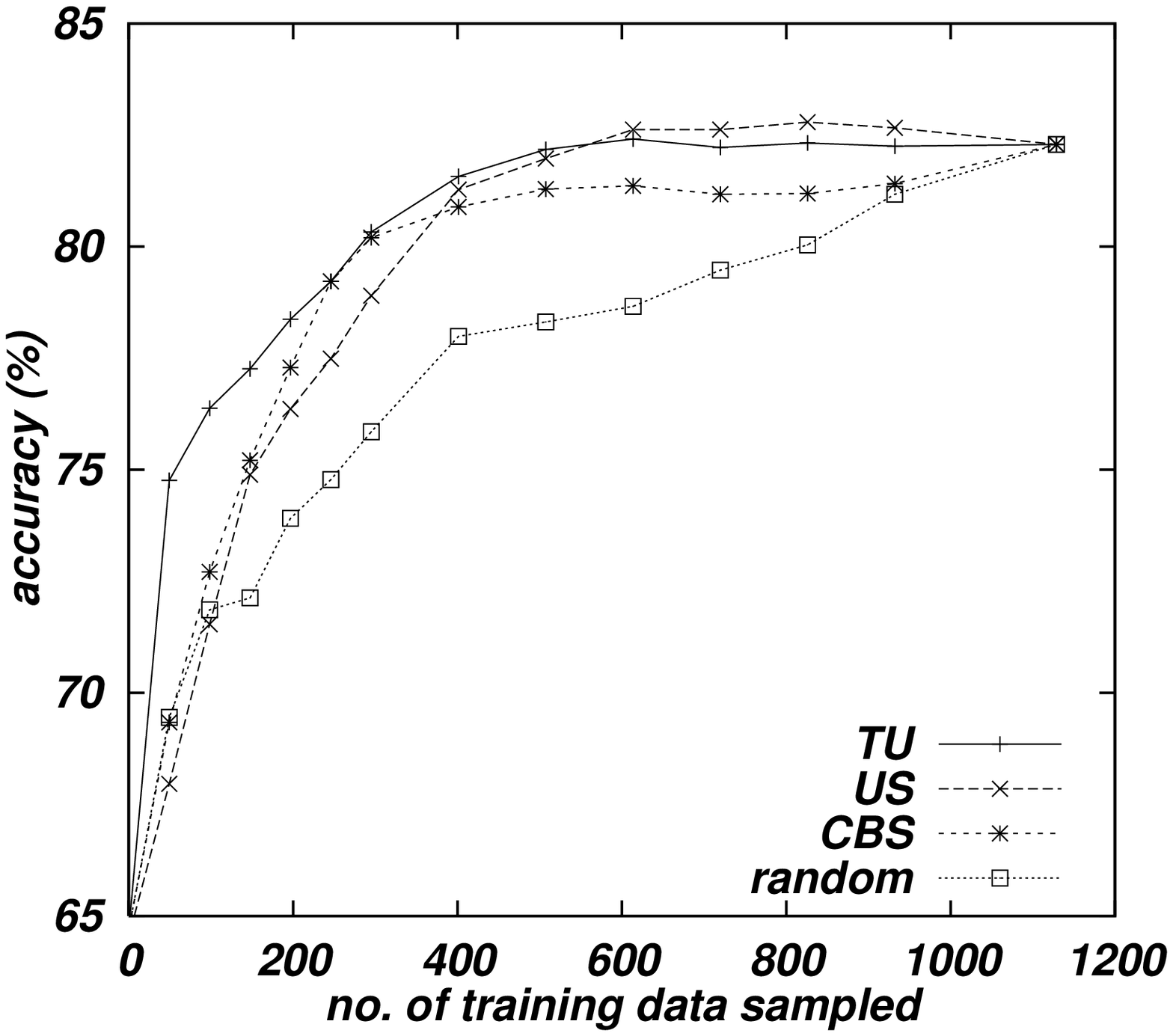,height=3.5in}}
  \medskip
  \caption{The relation between the number of training data sampled
    and accuracy of the system.}
  \label{fig:accuracy_IPAL}
\end{figure}

We evaluated each system performance according to its accuracy, that
is the ratio of the number of correct outputs, compared to the total
number of inputs.  For the purpose of this experiment, we set the
sample size to 1 for each iteration, \mbox{$\lambda = 0.5$} for
Equation~\eq{eq:certainty}, and \mbox{$k = 1$} for
Equation~\eq{eq:utility_temp}. Based on a preliminary experiment,
increasing the value of $k$ either did not improve the performance
over that for \mbox{$k = 1$}, or lowered the overall performance.
Figure~\ref{fig:accuracy_IPAL} shows the relation between the number
of the training data sampled and the accuracy of the system. In
Figure~\ref{fig:accuracy_IPAL}, zero on the x-axis represents the
system using only the examples provided by IPAL.  Looking at
Figure~\ref{fig:accuracy_IPAL} one can see that compared with random
sampling and committee-based sampling, our sampling method reduced the
number of the training data required to achieve any given
accuracy. For example, to achieve an accuracy of 80\%, the number of
the training data required for our method was roughly one-third of
that for random sampling.  Although the accuracy for our method was
surpassed by that for uncertainty sampling for larger sizes of
training data, this minimal difference for larger data sizes is
overshadowed by the considerable performance gain attained by our
method for smaller data sizes.

\begin{figure}
  \centering
  \mbox{\psfig{figure=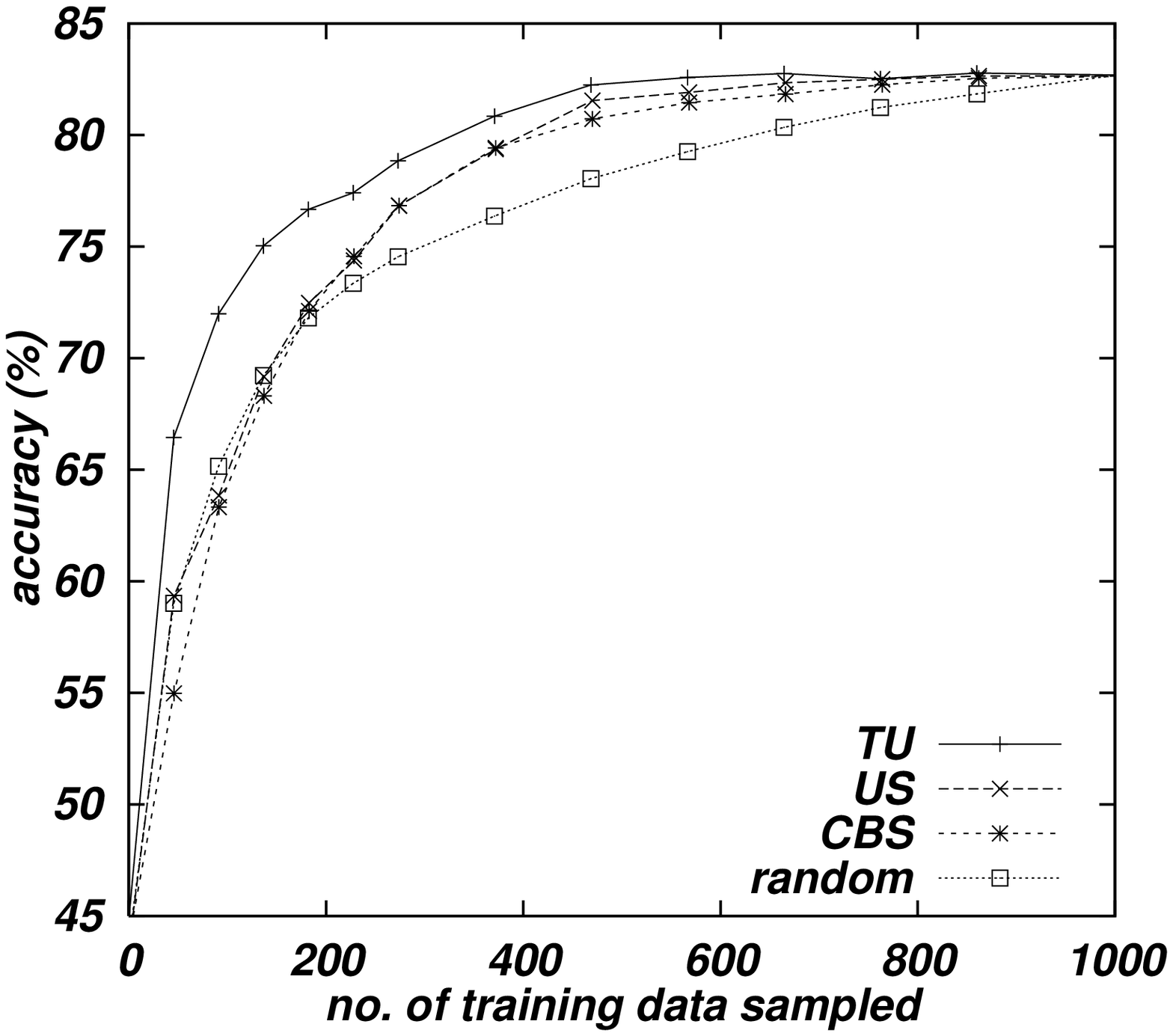,height=3.5in}}
  \medskip
  \caption{The relation between the number of training data sampled
    and accuracy of the system without using examples from IPAL.}
  \label{fig:accuracy}
\end{figure}

Since IPAL has, in a sense, been manually selectively sampled in an
attempt to model the maximum verb sense coverage, the performance of
each method is biased by the initial contents of the database. To
counter this effect, we also conducted an experiment involving the
construction of the database from scratch, without using examples from
IPAL. During the initial phase, the system randomly selected one
example for each verb sense from the training set, and a human expert
provided the correct interpretation to initialize the system.
Figure~\ref{fig:accuracy} shows the performance of the various
methods, from which the same general tendency as seen in
Figure~\ref{fig:accuracy_IPAL} is observable.  However, in this case,
our method was generally superior to other methods.  Through these
comparative experiments, we can conclude that our example sampling
method is able to decrease the number of the training data, i.e.,  the
overhead for both supervision and searching, without degrading the
system performance.

\subsection{Related Work}
\label{subsec:related}

\subsubsection{Uncertainty Sampling}
\label{subsubsec:lewis}

The procedure for uncertainty sampling~\cite{lewis:sigir-94} is as
follows, where \mbox{$C(x)$} represents the interpretation certainty
for an example $x$ (see our sampling procedure in
Section~\ref{subsec:overview} for comparison):
\begin{enumerate}
\item $WSD(\set{D}, \set{X})$
\item $e \leftarrow \arg\min_{x\in\set{X}}C(x)$ 
\item $\set{D} \leftarrow \set{D} \cup\{e\},~~\set{X}
  \leftarrow \set{X}\cap \overline{\{e\}}$
\item goto 1
\end{enumerate}

Let us discuss the theoretical difference between this and our method.
Considering Figure~\ref{fig:tu} again, one can see that the concept of
training utility is supported by the following properties:
  \renewcommand{\theenumi}{\alph{enumi}}
  \def\labelenumi{(\theenumi)}
\begin{enumerate}
\item an example which neighbors more unsupervised examples is more
  informative (Figure~\ref{fig:tu}(a)),
\item an example less similar to one already existing in the database
  is more informative (Figure~\ref{fig:tu}(b)).
\end{enumerate}
\renewcommand{\theenumi}{\arabic{enumi}} \def\labelenumi{\theenumi.}
Uncertainty sampling directly addresses the second property, but
ignores the first. It differs from our method more crucially when more
unsupervised examples remain, because these unsupervised examples have
a greater influence on the computation of training utility.  This can
also be seen in the comparative experiments in Section~\ref{sec:eval},
in which our method outperformed uncertainty sampling to the highest
degree in early stages.

\subsubsection{Committee-based Sampling}
\label{subsubsec:engelson}

In committee-based sampling~\cite{engelson:acl-96}, which follows the
``query by committee'' principle~\cite{seung:acm-ws-92}, the system
selects samples based on the degree of disagreement between models
randomly taken from a given training set (these models are called
``committee members'').  This is achieved by iteratively repeating the
steps given below, in which the number of committee members is given
as two without loss of generality:
\begin{enumerate}
\item draw two models randomly,
\item classify unsupervised example $x$ according to each model,
  producing classifications $C_1$ and $C_2$,
\item if $C_1 \neq C_2$ (the committee members disagree), select $x$
  for the training of the system.
\end{enumerate}

\begin{figure}
  \centering
  \mbox{\psfig{figure=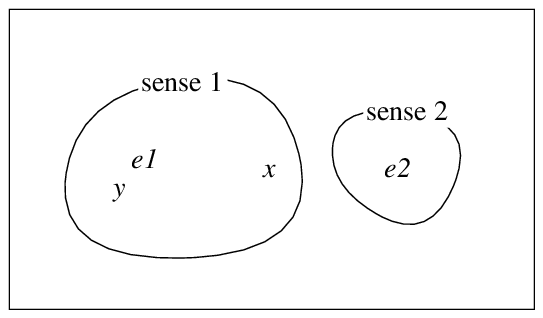,height=1.5in}}
  \medskip
  \caption{A case where either $x$ or $y$ can be selected in
    committee-based sampling.}
  \label{fig:cbs}
\end{figure}

Figure~\ref{fig:cbs} shows a typical disparity evident between
committee-based sampling and our sampling method. The basic notation
in this figure is the same as in Figure~\ref{fig:certainty}, and both
$x$ and $y$ denote unsupervised examples, or more formally
\mbox{$\set{D} = \{e_1,e_2\}$}, and \mbox{$\set{X} = \{x,y\}$}. Assume
a pair of committee members $\{e_1\}$ and $\{e_2\}$ have been selected
from the database $\set{D}$.  In this case, the committee members
disagree as to the interpretations of both $x$ and $y$, and
consequently, either example can potentially be selected as a sample
for the next iteration. In fact, committee-based sampling tends to
require a number of similar examples (similar to $e_1$ and $y$) in the
database, otherwise committee members taken from the database will
never agree. This is in contrast to our method, in which similar
examples are less informative. In our method, therefore, $x$ is
preferred to $y$ as a sample.  This contrast can also correlate to the
fact that committee-based sampling is currently applied to
statistics-based language models (HMM classifiers), in other words,
statistical models generally require that the distribution of the
training data reflects that of the overall text.  Through this
argument, one can assume that committee-based sampling is better
suited to statistics-based systems, while our method is more suitable
for example-based systems.

Engelson and Dagan~\shortcite{engelson:acl-96} criticized uncertainty
sampling~\cite{lewis:sigir-94}, which they call a ``single model''
approach, as distinct from their ``multiple model'' approach:
\begin{ext}
  sufficient statistics may yield an accurate 0.51 probability
  estimate for a class $c$ in a given example, making it certain that
  $c$ is the {\it appropriate\/} classification.\footnote{By
  appropriate classification, Engelson and Dagan mean the
  classification given by a perfectly trained model.} However, the
  certainty that $c$ is the {\it correct\/} classification is low,
  since there is a 0.49 chance that $c$ is the wrong class for the
  example. A single model can be used to estimate only the second type
  of uncertainty, which does not correlate directly with the utility
  of additional training. (p.325)
\end{ext}
We note that this criticism cannot be applied to our sampling method,
despite the fact that our method falls into the category of a single
model approach. In our sampling method, given sufficient statistics,
the increment of the certainty degree for unsupervised examples, i.e.,
the training utility of additional supervised examples, becomes small
(theoretically, for both example-based and statistics-based
systems). Thus, the utility factor can be considered to correlate
directly with additional training, for our method.

\section{Conclusion}
\label{sec:conclusion}

Corpus-based approaches have recently pointed the way to a promising
trend in word sense disambiguation. However, these approaches tend to
require a considerable overhead for supervision in constructing a
large-sized database, additionally resulting in a computational
overhead to search the database.  To overcome these problems, our
method, which is currently applied to an example-based verb sense
disambiguation system, selectively samples a smaller-sized subset from
a given example set.  This method is expected to be applicable to
other example-based systems. Applicability for other types of systems
needs to be further explored.

The process basically iterates through two phases: (normal) word sense
disambiguation and a training phase.  During the disambiguation phase,
the system is provided with sentences containing a polysemous verb,
and searches the database for the most semantically similar example to
the input (nearest neighbor resolution).  Thereafter, the verb is
disambiguated by superimposing the sense of the verb appearing in the
supervised example.  The similarity between the input and an example,
or more precisely the similarity between the case fillers included in
them, is computed based on an existing thesaurus. In the training
phase, a sample is then selected from the system outputs and provided
with the correct interpretation by a human expert.  Through these two
phases, the system iteratively accumulates supervised examples into
the database.  The critical issue in this process is to decide which
example should be selected as a sample in each iteration. To resolve
this problem, we considered the following properties: (a) an example
that neighbors more unsupervised examples is more influential for
subsequent training, and therefore more informative, and (b) since
our verb sense disambiguation is based on nearest neighbor resolution,
an example similar to one already existing in the database is
redundant.  Motivated by these properties, we introduced and
formalized the concept of training utility as the criterion for
example selection.  Our sampling method always gives preference to
that example which maximizes training utility.

We reported on the performance of our sampling method by way of
experiments in which we compared our method with random sampling,
uncertainty sampling~\cite{lewis:sigir-94} and committee-based
sampling~\cite{engelson:acl-96}. The result of the experiments showed that
our method reduced both the overhead for supervision and the overhead
for searching the database to a larger degree than any of the above
three methods, without degrading the performance of verb sense
disambiguation. Through the experiment and discussion, we claim that
uncertainty sampling considers property (b) mentioned above, but lacks
property (a).  We also claim that committee-based sampling differs
from our sampling method in terms of its suitability to
statistics-based systems as compared to example-based systems.

\starttwocolumn

\begin{acknowledgments}
  The authors would like to thank Manabu Okumura (JAIST, Japan),
  Timothy Baldwin (TITECH, Japan), Michael Zock (LIMSI, France), Dan
  Tufis (Romanian Academy, Romania) and anonymous reviewers for their
  comments on an earlier version of this paper. This research is
  partially supported by a Research Fellowship of the Japan Society
  for the Promotion of Science for Young Scientists.
\end{acknowledgments}

\end{document}